\documentclass{article}
\usepackage{style,amsmath,graphicx,booktabs,multirow,amssymb}

\usepackage{subcaption}
\usepackage{cite}
\usepackage{xurl}

\makeatletter
\newcommand{\authoraddress}[1]{\gdef\@authoraddress{#1}}
\let\@authoraddress\@empty
\let\origmaketitle\maketitle

\renewcommand{\maketitle}{%
  \name{\@author}%
  \address{\@authoraddress}%
  \origmaketitle
}
\makeatother

\title{Rule-Constrained Assignment for Cue-Ball Identification\\in Broadcast Snooker}

\author{
Yuxin Cao$^{1}$ \ \ Wei Song$^{2}$ \ \ Yuezhong Wu$^{3}$ \ \ Jin Song Dong$^{1}$
}
\authoraddress{
$^1$National University of Singapore, Singapore\\$^2$Griffith University, Australia\\$^3$Fuzhou University, China
}

\begin{document}
\maketitle

\begin{abstract}
Accurate cue-ball identification is essential for metric analysis of broadcast snooker. Existing systems evaluate each candidate independently against a fixed white prototype and reject candidates above an appearance threshold. Under broadcast conditions, illumination changes can make colored balls appear white, while intrusions from players and equipment can obscure the cue ball or introduce competing candidates. We formulate cue-ball identification as a rule-constrained assignment problem that jointly assigns detected candidates to the bounded snooker inventory: one cue ball, up to 15 reds, and six colors with known spots. The cue ball is selected by the incremental cost of assigning each candidate to the white slot, and the conventional appearance test follows as the one-slot case. On 419 hand-annotated shots, our method improves identity accuracy from 88.1\% to 95.5\%, and from 80.5\% to 95.2\% on held-out venues. Within CueLift, our metric state-recovery system, the assignment expands coverage from 36.2\% to 55.4\% over 6,241 scorable shots. Assigning an estimate to every shot in a separate evaluation on 2,529 shots preserves this advantage.
\end{abstract}


\begin{figure*}[t]
\centering
\includegraphics[width=0.95\textwidth]{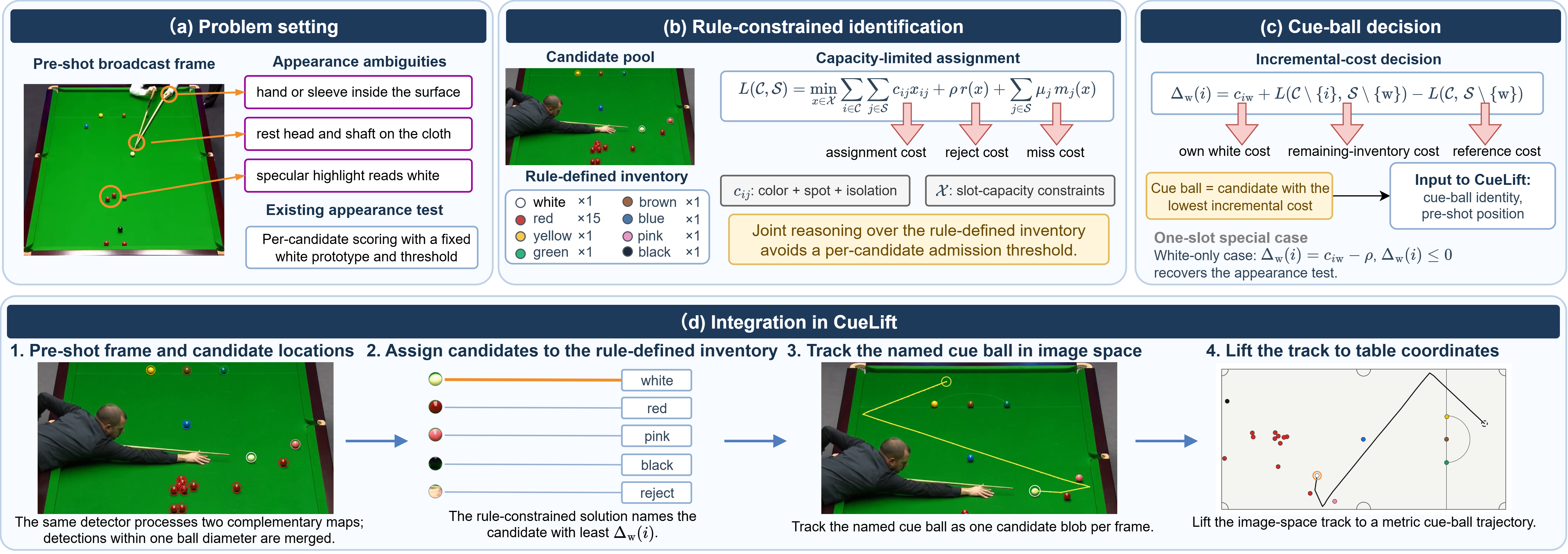}
\caption{Overview of the proposed rule-constrained cue-ball identification and its integration in CueLift.}
\label{fig:overview}
\end{figure*}

\section{Introduction}
\label{sec:intro}

A regulation snooker table has fixed dimensions and contains up to 22 balls of known radius and color. This structure makes metric state recovery from broadcast footage possible without instrumenting the table or altering the production pipeline. A prerequisite for such recovery is identifying which image location corresponds to the cue ball. Fig.~\ref{fig:overview}(a) illustrates the principal ambiguities: a hand or sleeve may enter the playing surface, the head or shaft of a rest may lie on the cloth, and a specular highlight may make a colored ball appear white. Venue-dependent illumination can also leave the cue ball darker than surrounding candidates. These effects can either conceal the cue ball or introduce a more plausible false candidate, causing an identification error that propagates into tracking and metric reconstruction.

Prior vision systems for billiards and snooker parse broadcast shots and recover table states or trajectories \cite{denman2003,chou2009,guo2007,jiang2013}, whereas the public trajectory benchmark uses a fixed overhead camera \cite{rodriguez2023}. They classify balls primarily from color or luminance \cite{denman2003,guo2007,jiang2013}. Recent research has focused on single-image analysis and classification \cite{pix2pockets,attentionpool,billiardphys,zhang2024}. Monocular state recovery is more developed in table tennis \cite{tt3d,tt4d}, where apparent ball diameter provides a depth cue \cite{vanzandycke2022}. Sports trackers infer identity from appearance and motion \cite{tracknet,tracknetv4}, sometimes using detection thresholds tuned per sequence \cite{ghost2023}. However, these candidate-wise methods ignore information fixed before observing the image: snooker defines a bounded inventory of one cue ball, up to 15 reds, and one ball of each of six colors. Although exclusion constraints have been studied in multi-object tracking \cite{maccormick2000}, they have not been applied to joint assignment for broadcast cue-ball identification.

Our core insight is that cue-ball identification is inherently joint because candidates compete for fixed identities. A bright false candidate may resemble the cue ball in isolation but better fit a colored slot through its color, position, and surroundings, leaving a dim or occluded cue ball as the more consistent choice. Evidence for each candidate therefore depends on both its white appearance and how its removal changes the optimal assignment of the remaining candidates. We jointly assign all candidates to rule-defined slots and select the cue ball by the objective increase from assigning each candidate to the white slot. With only the white slot retained, the formulation reduces to the conventional appearance test.

The main contribution of this paper is a rule-constrained formulation of cue-ball identification that replaces independent appearance-based admission with joint assignment over the bounded snooker inventory. Direct evaluation on 419 hand-annotated shots shows an accuracy improvement from 88.1\% to 95.5\%; on held-out venues, the gain is from 80.5\% to 95.2\%. We also introduce CueLift, a metric state-recovery system for broadcast snooker, to evaluate the method within a complete pipeline. With the proposed assignment, CueLift covers 55.4\% of 6,241 scorable shots, compared with 36.2\% using the appearance test. A separate evaluation assigning an estimate to every shot preserves the advantage on 2,529 shots.

\section{Method}
\label{sec:method}

Given a pre-shot broadcast frame, our method constructs a candidate pool, assigns its elements jointly to the rule-defined snooker inventory, and selects the candidate with the lowest incremental white-slot cost. Fig.~\ref{fig:overview} summarizes the formulation and its integration into CueLift.

\noindent\textbf{Candidate Pool.} As part of CueLift, we generate candidate locations by applying the same matched-filter detector to two complementary maps of the pre-shot frame. On the cloth-departure map, it detects the colored balls and most cue balls. On a second map with nonwhite regions suppressed, it recovers cue balls that the first pass may miss in favor of brighter neighboring candidates. Detections separated by less than one ball diameter are merged into a single candidate.

\noindent\textbf{Rule-Defined Inventory and Capacity-Limited Assignment.} Let $\mathcal{C}$ denote the candidate pool and $\mathcal{S}$ the set of slots defined by the rules of snooker: one white slot, one red slot, and one slot for each of the six colors. Slot $j$ accepts at most $b_j$ candidates, where $b_{\mathrm{r}}=15$ for the red slot and $b_j=1$ for every other slot. Let $x_{ij}\in\{0,1\}$ indicate whether candidate $i$ is assigned to slot $j$. We define $r(x)=|\mathcal{C}|-\sum_{ij}x_{ij}$ as the number of rejected candidates and $m_j(x)=b_j-\sum_i x_{ij}$ as the unfilled capacity of slot $j$. The capacity-limited assignment is then obtained by solving
\begin{equation}
\label{eq:assignment}
L(\mathcal{C},\mathcal{S})=\min_{x\in\mathcal{X}}\sum_{i\in\mathcal{C}}\sum_{j\in\mathcal{S}}c_{ij}x_{ij}+\rho\,r(x)+\sum_{j\in\mathcal{S}}\mu_j\,m_j(x),
\end{equation}
where $\mathcal{X}$ contains assignments in which each candidate occupies at most one slot and no slot exceeds its capacity. The reject cost $\rho$ penalizes each unassigned candidate, while $\mu_j$ penalizes unfilled capacity in slot $j$. We set $\mu_{\mathrm{r}}=0$ because occlusion may leave fewer than 15 reds visible. The miss costs on the six color slots encourage plausible colors to be represented in the assignment. Without them, the optimizer may leave these slots empty, weakening the inventory information used to separate competing candidates.

\noindent\textbf{Assignment Costs.} The cost $c_{ij}$ of assigning candidate $i$ to slot $j$ contains color, spot, and isolation terms. The color term measures the weighted Lab distance to the corresponding class prototype. We compute this distance over both the interior and the full extent of the candidate and retain the larger value, preventing a specular highlight on a colored ball from dominating its assignment to the white slot. The spot term favors assigning each color near its rule-defined spot. Its distance is capped at four ball radii so that a color away from its spot is penalized rather than excluded. The isolation term penalizes candidates whose surrounding region departs from the cloth as strongly as the candidate center, a pattern characteristic of hands, sleeves, and rests.

\noindent\textbf{Incremental-Cost Decision.} Let $\mathrm{w}$ denote the white slot. For each candidate $i$, we measure the incremental cost of assigning it to $\mathrm{w}$ relative to the optimal assignment without $\mathrm{w}$:
\begin{equation}
\label{eq:incremental-cost}
\Delta_{\mathrm{w}}(i)=c_{i\mathrm{w}}+L(\mathcal{C}\setminus\{i\},\,\mathcal{S}\setminus\{\mathrm{w}\})-L(\mathcal{C},\,\mathcal{S}\setminus\{\mathrm{w}\}).
\end{equation}
We identify the cue ball as $\hat{\imath}=\arg\min_{i\in\mathcal{C}}\Delta_{\mathrm{w}}(i)$, breaking ties using the lower white-slot cost. The first term measures the candidate's compatibility with the white slot, while the remaining terms measure how its removal affects the optimal assignment of all other candidates. Because both assignment terms exclude $\mathrm{w}$, the white-slot miss cost does not enter this decision. This joint comparison can identify a dim cue ball partially occluded by a bridge hand even when another candidate appears whiter. Both optimization problems in \eqref{eq:incremental-cost} are capacitated linear assignments. Minimizing over the candidate pool makes the method refusal-free when the pool is nonempty, although it can fail if the cue ball is absent or another candidate has a lower incremental cost.

\noindent\textbf{Appearance Test as the One-Slot Case.} When only $\mathrm{w}$ is retained, $L(\mathcal{C},\emptyset)=\rho|\mathcal{C}|$, and \eqref{eq:incremental-cost} reduces to $\Delta_{\mathrm{w}}(i)=c_{i\mathrm{w}}-\rho$. Therefore, $\Delta_{\mathrm{w}}(i)\leq 0$ exactly when $c_{i\mathrm{w}}\leq\rho$. With the same white-appearance cost and $\rho$ set to the baseline threshold, this condition recovers the conventional appearance test. The baseline is thus the one-slot form of our assignment with an additional admission decision, whereas the full-inventory formulation selects a candidate whenever the pool is nonempty.

\noindent\textbf{Integration in CueLift.} CueLift segments broadcast footage into shots, removes players from the playing surface, and registers a three-dimensional regulation table model to the camera \cite{tvcalib,pnlcalib}. The candidate selected by the assignment initializes cue-ball tracking in image space. CueLift then maps the tracked center into table coordinates, using the geometric constraint that the center of a ball on the cloth lies one radius above the table surface.

\section{Experiments}
\label{sec:exp}

\subsection{Experimental Setup}

We evaluate cue-ball identification directly and assess its effect on end-to-end metric state recovery within CueLift.

\noindent\textbf{Benchmark and Split.} Our benchmark contains 7,501 shots from 248 matches at 21 venues, drawn from 343 videos and 541.0 h of televised snooker published by World Snooker Tour \cite{wst} and Matchroom \cite{matchroom}. A venue-level split prevents any venue, match, staging, or shot from appearing on both sides, yielding 11 training and 10 held-out venues.

\noindent\textbf{Implementation Details.} Class prototypes are extracted from a rack frame for 121 of the 248 matches, with a fixed built-in set used for the remainder. The reject and miss costs and two geometric weights are fixed according to the scales of their terms, while the isolation weight is selected on the training venues. No parameter is tuned on held-out venues or fitted per venue. Each optimization in \eqref{eq:incremental-cost} takes approximately 1 ms per frame, and the table registration localizes resting balls to 11.3 mm.

\noindent\textbf{Protocol and Metrics.} Each shot ends with the balls at rest, allowing the subsequent still frame to provide a reference without per-frame trajectory annotation. We withhold this frame from every method and use it only to evaluate the predicted resting position. A shot is scorable when the frame provides a resting cue-ball position, which holds for 6,241 shots. Coverage is measured over this set.

Since the reference is detected automatically, it names the wrong object on 10.5\% of hand-annotated shots. We therefore apply an independent chromatic screen to the withheld frame. It accepts 5,323 references, rejecting 25.5\% on held-out venues and 4.4\% on training venues. Resting error is evaluated on the accepted references using the median distance in millimeters and conditional success within 200 mm.

\noindent\textbf{Identity Annotations.} Direct evaluation prevents downstream reacquisition from concealing an incorrect initial identity. An annotator marked the cue-ball center in magnified crops while blind to both methods' outputs, and each mark was matched to the nearest candidate. The first set contains 419 shots answered by both identification methods. The second contains 150 shots sampled from the 985 shots with accepted references for which the appearance-test pipeline returns no trajectory while the proposed pipeline does. Reannotation agrees on 29 of 30 shots.

\noindent\textbf{Baselines.} We compare our method with the one-slot appearance test, which follows the appearance-based design used in prior vision systems for billiards and snooker \cite{denman2003,guo2007,jiang2013}. It uses the same white-appearance cost and a threshold selected on the training venues. The classical pipeline thresholds the cloth in HSV, maps the table to a regulation rectangle using a homography, detects connected components, and classifies them using fixed color thresholds.

\subsection{Results and Analysis}

\noindent\textbf{Cue-Ball Identification.} Table~\ref{tab:naming} reports identity accuracy against hand annotations, counting a refusal as incorrect. On all 419 shots, our method reaches 95.5\%, compared with 88.1\% for the appearance test. On held-out venues, the improvement is larger, from 80.5\% to 95.2\%. Our method is correct on 36 shots that the appearance test gets wrong, while the reverse occurs on five. An exact McNemar test over the 41 discordant pairs gives $p=7.8\times10^{-7}$.

\begin{figure*}[t]
\centering
\includegraphics[width=0.92\textwidth]{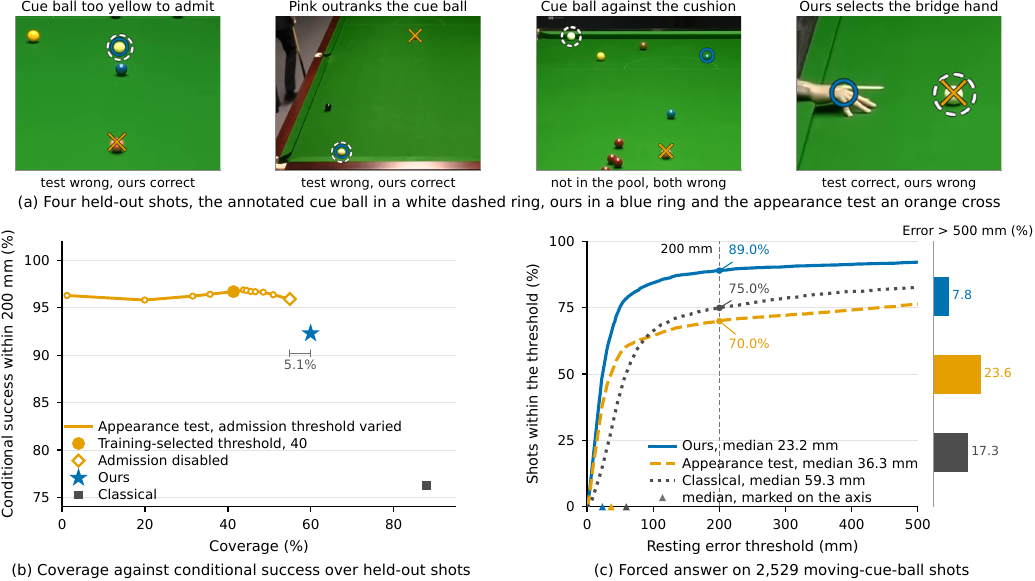}
\caption{Qualitative and end-to-end evaluation with annotated cases, threshold variation, and forced answers.}
\label{fig:results}
\end{figure*}

On the 150 newly covered shots, our method reaches 76.7\% compared with 38.0\% for the appearance test. The appearance test returns an identity on only 65 shots and is correct on 57. The 80 discordant outcomes favor our method by 69 to 11, giving $p=2.1\times10^{-11}$. The cue ball is absent from the candidate pool in 12 shots, limiting any method using this pool to 92.0\%. Fig.~\ref{fig:results}(a) shows four held-out cases selected from predefined categories: two corrected appearance-test errors, one failure shared by both methods, and one case in which the appearance test is correct but the proposed assignment selects the bridge hand.

\begin{table}[t]
\caption{Cue-ball identity accuracy against hand annotations. A refusal counts as incorrect.}
\label{tab:naming}
\centering
\footnotesize
\begin{tabular}{@{}lrrr@{}}
\toprule
Method & All 419 & Held out & Newly covered 150 \\
\midrule
The appearance test & 88.1\% & 80.5\% & 38.0\% \\
Our method & \textbf{95.5\%} & \textbf{95.2\%} & \textbf{76.7\%} \\
\bottomrule
\end{tabular}
\vspace{-2mm}
\end{table}

\noindent\textbf{End-to-End Recovery.} Across all 6,241 scorable shots, CueLift with the proposed assignment achieves 55.4\% coverage, compared with 36.2\% when using the appearance test. On all 3,054 held-out scorable shots, coverage increases from 38.9\% to 59.4\%. On the 5,323 shots with accepted references, our system achieves 55.0\% coverage at a median error of 22.1 mm and 93.5\% conditional success within 200 mm. Fig.~\ref{fig:results}(b) reports the corresponding results on the 2,275 held-out shots retained by the reference screen. Our system achieves 60.0\% coverage, a median error of 24.8 mm, and 92.3\% conditional success within 200 mm. The appearance-test baseline achieves 41.4\%, 22.6 mm, and 96.7\%, while the classical pipeline achieves 87.9\%, 71.2 mm, and 76.2\%. Across 1,196 paired trajectories, our system has 41.7 mm lower error than the classical pipeline and is nearer on 81.4\% of shots. Missing trajectories arise from candidate-generation or tracking failures rather than abstention by the assignment.

\noindent\textbf{Admission Threshold.} Fig.~\ref{fig:results}(b) evaluates the appearance test at 11 admission thresholds on the same 2,275 held-out shots. Increasing the threshold from the training-selected value of 40 to 90 raises coverage from 41.4\% to 50.9\%, while conditional success remains between 95.8\% and 96.9\%. Disabling admission raises coverage to 54.9\% with 95.9\% conditional success, compared with 60.0\% and 92.3\% for CueLift with the proposed assignment. Threshold variation therefore does not reproduce the coverage attained by the assignment. Threshold variation is retrospective, whereas other results use the training-selected value.

\noindent\textbf{Forced Answer.} Selective abstention can confound conditional accuracy, so we assign an estimate to every one of the 2,529 moving-cue-ball shots, defined by cue-ball travel of at least 100 mm. Each method returns its recovered trajectory when available and otherwise uses the benchmark's detected pre-shot cue-ball location as its resting-position estimate. Under this protocol, our system achieves a median error of 23.2 mm with 89.0\% of shots within 200 mm, compared with 36.3 mm and 70.0\% for the appearance-test baseline and 59.3 mm and 75.0\% for the classical pipeline. Using the pre-shot location alone gives an error within 200 mm on 7.3\% of these shots. Fig.~\ref{fig:results}(c) shows the success curves up to 500 mm and the errors beyond this range.

\noindent\textbf{Sensitivity and Scope.} We evaluate the five constants over 130 one-at-a-time configurations, ranging from disabling each term to making it dominant, and 72 Latin hypercube samples that vary all five jointly. Held-out coverage changes by at most 1.4\% relative to its reported value, and pooled median error by at most 1.1\%. The results therefore do not depend strongly on the precise parameter values. The remaining coverage bottleneck is tracking, which returns fewer than three observations on 2,429 of the 6,241 scorable shots.

\noindent\textbf{Limitations.} Both annotation sets were produced by a single annotator without independent adjudication, although the annotator was blind to the cue-ball identities predicted by all evaluated methods. The benchmark includes only shots whose two still frames contain no players, and videos from one broadcast producer appear in both partitions. Future work will use the retained annotations for independent re-adjudication, include shots with player occlusion, and evaluate on splits fully disjoint by broadcast producer.

\section{Conclusion}
\label{sec:conclusion}

In this paper, we show that cue-ball identity can be inferred by jointly assigning detected candidates under the inventory constraints imposed by the rules of snooker. This rule-constrained formulation replaces independent appearance-based admission and improves held-out identity accuracy from 80.5\% to 95.2\%. We also introduce CueLift, a metric state-recovery system in which the proposed method increases coverage from 36.2\% to 55.4\%. The advantage remains when every shot is assigned an estimate in the evaluation on the 2,529 shots where the cue ball moved.

\clearpage
\bibliographystyle{IEEEbib}
\bibliography{refs}

\end{document}